\documentclass[journal]{IEEEtran}
\usepackage{cite}
\usepackage{amsmath,amssymb}
\usepackage{array}
\usepackage{graphicx}
\usepackage{wrapfig}
\usepackage{xparse}
\usepackage{xkeyval}
\usepackage{stfloats}
\usepackage{url}
\usepackage[T1]{fontenc}
\usepackage[table,dvipsnames,svgnames,x11names]{xcolor}
\usepackage{tikz}
\usepackage{booktabs}
\usepackage{multirow}
\usepackage{colortbl}
\usepackage{tabularx}
\usepackage[colorlinks,linkcolor=blue,filecolor=blue,urlcolor=blue,citecolor=blue]{hyperref}

\newwrite\authorbibfile
\AtBeginDocument{%
  \immediate\openout\authorbibfile=\jobname.aub%
}
\AtEndDocument{%
  \immediate\closeout\authorbibfile
  \InputIfFileExists{\jobname.aub}{}{}
}

\makeatletter
\define@key{authorbib}{scale}[1]{%
  \def\AuthorbibKVMacroScale{#1}%
}
\define@key{authorbib}{wraplines}[10]{%
  \def\AuthorbibKVMacroWraplines{#1}%
}
\define@key{authorbib}{imagewidth}[4cm]{%
  \def\AuthorbibKVMacroImagewidth{#1}%
}
\define@key{authorbib}{overhang}[10pt]{%
  \def\AuthorbibKVMacroOverhang{#1}%
}
\define@key{authorbib}{imagepos}[l]{%
  \def\AuthorbibKVMacroImagepos{#1}%
}
\makeatother

\presetkeys{authorbib}{imagepos=l, imagewidth=4cm, wraplines=8, overhang=20pt}{}
\newlength{\AuthorbibTopSkip}
\newlength{\AuthorbibBottomSkip}
\NewDocumentCommand{\authorbibliography}{+o+m+m}{%
  \IfNoValueTF{#1}{%
  }{%
    \setkeys{authorbib}{#1}%
    \immediate\write\authorbibfile{%
      \string\begin{wrapfigure}[\AuthorbibKVMacroWraplines]{\AuthorbibKVMacroImagepos}[\AuthorbibKVMacroOverhang]{\AuthorbibKVMacroImagewidth}^^J
        \string\includegraphics[scale=\AuthorbibKVMacroScale]{#2}^^J
      \string\end{wrapfigure}^^J
    }%
  }%
  \immediate\write\authorbibfile{%
    \unexpanded{\vspace{\AuthorbibTopSkip}}^^J
    \string\noindent\relax
    \unexpanded{\textbf{#3}\par}^^J
    \unexpanded{\vspace{\AuthorbibBottomSkip}}^^J
  }%
}%

\begin{document}
\title{THG-Omni: Traceable Human-Centric Graph Supervision for Video Reasoning}

\author{Jisheng Dang, Kai Wei, Ronghao Lin, Bimei Wang, Hong Peng, Wei-Shi Zheng, 
   ~\IEEEmembership{Senior Member, ~IEEE},
 Bin Hu, ~\IEEEmembership{Fellow, ~IEEE}, Qi Tian, ~\IEEEmembership{Fellow, ~IEEE}, Tat-Seng Chua
\thanks{Jisheng Dang, Kai Wei, Bimei Wang, and Hong Peng are with School of Information Science \& Engineering, Lanzhou University, Lanzhou 730000, China.  (E-mail: dangjsh@mail2.sysu.edu.cn).}
 \thanks{Wei-Shi Zheng is with the School of Computer Science and Engineering, Sun Yat-sen University, Guangzhou 510006, China (E-mail: wszheng@ieee.org).}
\thanks{Bin Hu is with the School of Medical Technology, Beijing Institute of Technology, Beijing 100081, China (E-mail: bh@bit.edu.cn). }
\thanks{Qi Tian is with Cloud and AI BU, Huawei, Shenzhen, Guangdong 518129,
China (e-mail: tian.qi1@huawei.com).}
\thanks{Tat-Seng Chua is with the School of Computing, National University of Singapore, Singapore 119077 (E-mail: dcscts@nus.edu.sg). }
\thanks{*Corresponding authors: Hong Peng and Bin Hu.}}

\maketitle

\begin{abstract}
Omni-modal models can answer questions about video, but their predictions rarely show which participant, event, or interval supports an answer. Moreover, answer-level supervision alone does not teach a model to interpret structured video evidence. To systematically mitigate this issue, we present THG-Omni, a human-centric graph-supervision framework that links Segment Anything Model~3 participant tracks with temporally grounded Qwen3-VL annotations. Candidate semantic claims are checked against the source video and available audio before they are used to construct training questions, while track geometry provides explicit spatial and temporal supervision. The resulting questions teach an omni-modal reasoning model to interpret participant attributes, actions, affect, interactions, and their timing from a structured graph. At inference, THG-Omni uses a predicted graph constructed from the evaluation video. Experiments on IntentBench show consistent improvements over the unadapted backbone across social reasoning, emotion understanding, and deception perception. On NExT-QA, THG-Omni achieves 89.54\% accuracy with the matched graph, compared with 85.10\% without graph input. Further validation experiments show that compact graph input achieves 78.22\% accuracy, compared with 78.02\% using the full graph and 77.58\% without graph input. A complementary diagnostic demonstrates that the predicted graph preserves evidence useful for answering questions even without the source video and audio. Together, these findings show that THG-Omni improves human-centric video reasoning while grounding its evidence in identifiable participants and events. Our codes are available at \url{https://anonymous.4open.science/r/review-n94hhfcwp5666/}
\end{abstract}


\section{Introduction}\label{sec:introduction}

\newcommand{\rdzero}{\tikz[baseline=-0.6ex]\node[circle,draw=black!25,fill=black!10,inner sep=2.0pt]{};}
\newcommand{\rdhalf}{\tikz[baseline=-0.6ex]\node[circle,draw=Orange!70,fill=Orange!45,inner sep=2.0pt]{};}
\newcommand{\rdone}{\tikz[baseline=-0.6ex]\node[circle,draw=ForestGreen!80,fill=ForestGreen!55,inner sep=2.0pt]{};}

\newcommand{\learnpanel}[6][8]{%
\begin{tikzpicture}[font=\small]
    \node[
        draw=#6!70,
        fill=#6!#1,
        rounded corners=2pt,
        minimum width=0.225\textwidth,
        minimum height=2.45cm
    ] (box) {};

    \node[font=\bfseries] at ([yshift=0.82cm]box.center) {#2};
    \node[align=center] at ([yshift=0.28cm]box.center) {#3};
    \node[font=\scriptsize, align=center] at ([yshift=-0.32cm]box.center) {#4};
    \node[
        draw=#6!70,
        fill=white,
        rounded corners=1pt,
        font=\scriptsize,
        inner sep=3pt
    ] at ([yshift=-0.88cm]box.center) {#5};
\end{tikzpicture}%
}

Recent advances in multimodal large language models (MLLMs) have improved temporal localization, long-context modeling, and question answering for complex videos
\cite{ren2024timechat,wu2024longvideobench,zeng2025timesuite}.
By sampling more frames and retaining more visual tokens, these models can process longer videos and capture a wider range of events. However, increased visual coverage does not ensure that relevant evidence is correctly localized, associated with the proper participant, or used during inference
\cite{wu2024longvideobench,chen2025cg,li2025tigv}.
These limitations are particularly important for human-centric video reasoning, where a prediction may depend on brief and distributed evidence from facial expressions, body motion, speech, vocal delivery, and interactions among participants
\cite{cheng2024emotion,yang2025humanomniv2}.

Human-centric video reasoning requires persistent participant identity and explicit temporal association. Visible attributes, actions, and observable affective states must be assigned to the correct person and restricted to the interval in which they occur. Occlusion, camera motion, viewpoint changes, and shot transitions make such associations difficult
\cite{ravi2025sam2,gao2025multiple}.
Promptable video segmentation provides a useful basis for addressing this problem. Segment Anything Model~3 (SAM~3) grounds concept prompts and propagates instance masks over time, producing participant trajectories with explicit temporal and spatial information
\cite{carion2026sam}.
Video scene graphs further organize trajectories into participant nodes, temporal states, and relations
\cite{fei2024finsta,xiao2023covgt,kim2025weakly,huang2025laser}.
Related studies on multimodal knowledge graphs and structured representations also show that explicit entities and relations can support multimodal reasoning
\cite{lee2024multimodal,zhang2024native,huang2024structure}.

\begin{figure*}[t]
\centering
\includegraphics[width=\textwidth]{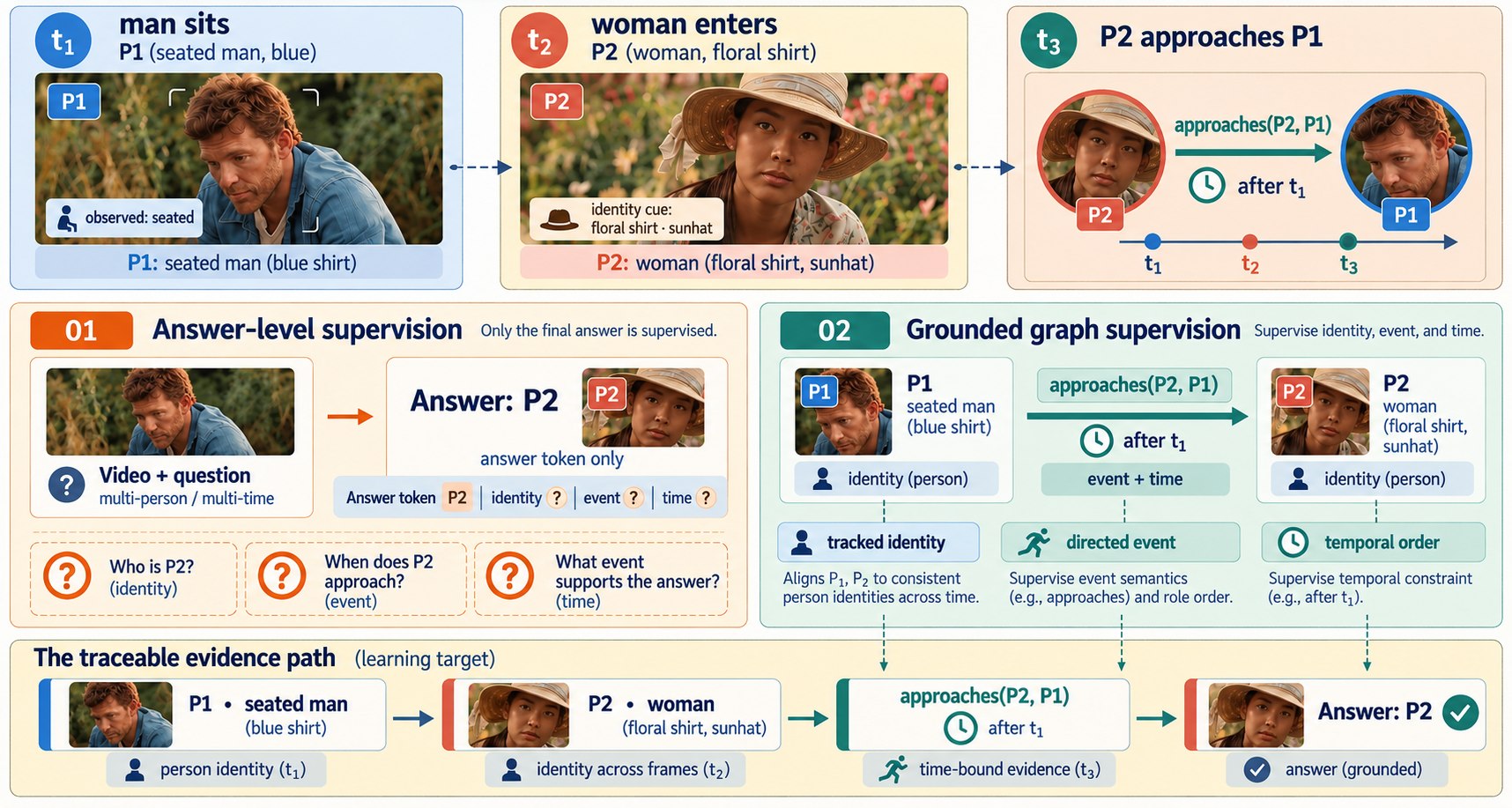}
\caption{Motivation for traceable graph supervision. This conceptual example contrasts answer-level supervision with identity- and time-grounded graph supervision. The depicted participants and event illustrate the learning objective rather than a benchmark prediction.}
\label{fig:motivation}
\end{figure*}

Nevertheless, constructing a human-centric video graph does not guarantee improved video reasoning. First, tracking or annotation errors can attach a valid observation to the wrong participant or extend it beyond its supported interval. Such errors may remain linguistically plausible and cannot be detected through schema validation alone. Second, a pretrained MLLM may not understand a serialized graph without targeted supervision. Direct graph injection can duplicate visual information, propagate uncertain associations, or receive little attention from a model that has not learned the meaning of the graph fields. Modality priors and visual attention biases may further prevent MLLMs from using supplied evidence
\cite{zhou2025mitigating,kang2025see}. 
Modular video question answering methods have also
explored intermediate evidence and decomposed
reasoning~\cite{min2024morevqa}. Therefore, graph construction and graph utilization must be treated as separate problems (Fig.~\ref{fig:motivation}).

\begin{figure*}[t]
    \centering
    \includegraphics[width=0.98\textwidth]{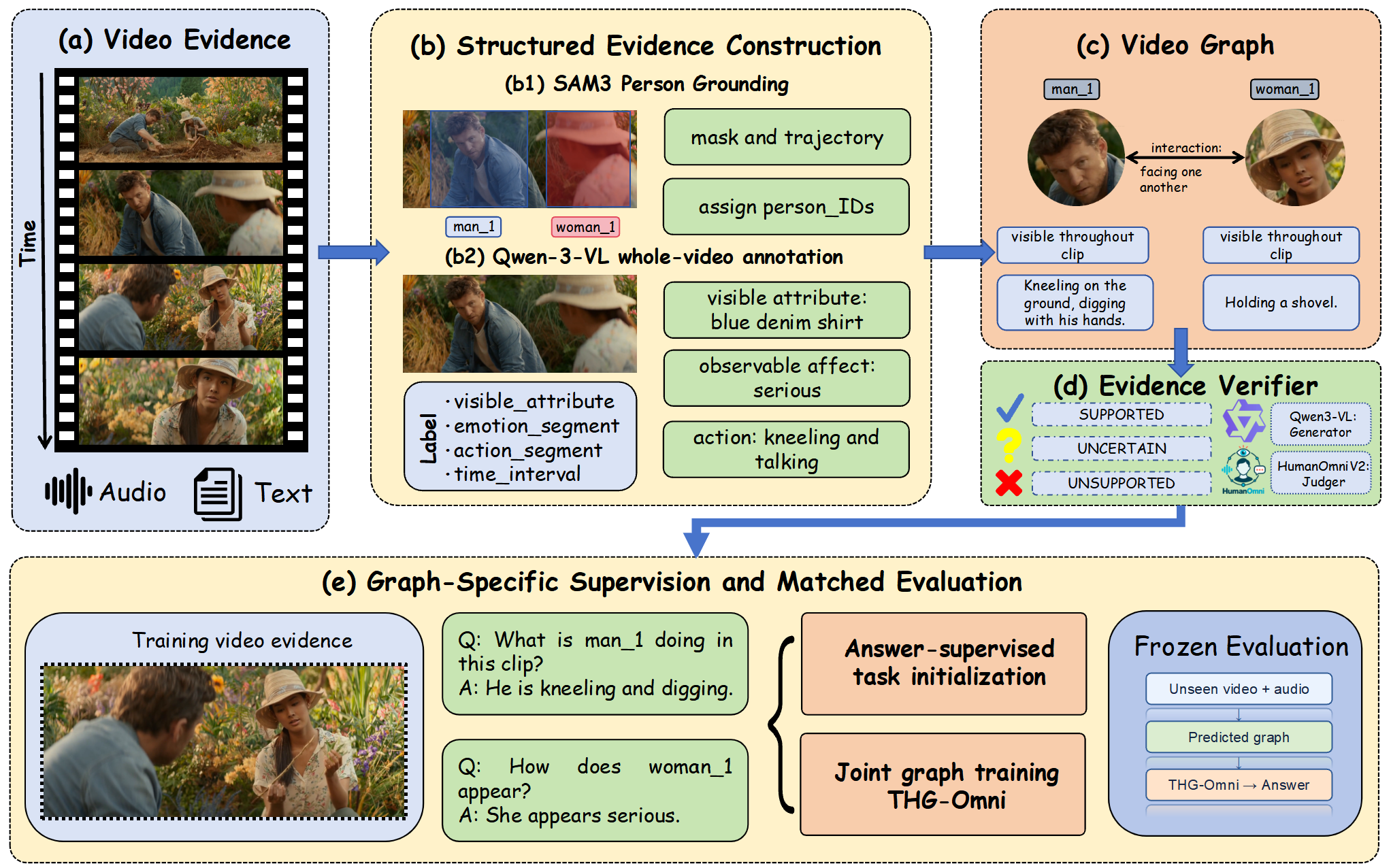}
    \caption{
    Overview of the proposed traceable human-centric graph-supervision framework.
    (a) Temporally ordered video evidence.
    (b) SAM3 grounds and tracks participants, while Qwen3-VL annotates visible attributes, emotions, actions, interactions, and their temporal intervals.
    (c) The grounded evidence forms a video graph with identity-bound nodes, temporal states, and participant relations.
    (d) HumanOmniV2 verifies each candidate claim using the source video and available audio; only supported claims are retained.
    (e) The verified graph supports training of THG-Omni, which is evaluated with video-matched, shuffled, and omitted graph input.
    }
    \label{fig:framework}
\end{figure*}

To address these limitations, we propose THG-Omni, a traceable human-centric graph-supervision framework for omni-modal video reasoning. As illustrated in Fig.~\ref{fig:framework}, SAM~3 first grounds and tracks question-relevant participants, providing visibility intervals, trajectories, and spatial relations. Qwen3-VL then generates whole-video semantic claims describing visible attributes, actions, observable affective states, interactions, and intention hypotheses. These geometric and semantic elements are merged into a human-centric video graph in which every claim is associated with a participant and a temporal interval. Unlike methods that only append a graph to the input prompt, our framework converts individual graph fields into explicit supervision examples.

Because model-generated semantic claims cannot be regarded as ground truth, we introduce cross-model verification before graph supervision. HumanOmniV2 evaluates each semantic claim using the source video and available audio, while human quality control examines participant identity, temporal extent, and visual support
\cite{chen2024mllm,yang2025humanomniv2}.
Only supported positive claims are retained. This protocol avoids synthetic negative statements that may introduce lexical shortcuts or semantically ambiguous counterfactuals. The verified semantic claims are combined with deterministic questions derived from participant trajectories to supervise graph interpretation. Using low-rank adaptation
\cite{hu2021lora}, we evaluate graph-specific supervision against the frozen model and compare video-matched, shuffled, and omitted graph input within checkpoint-aligned runs. These experiments examine information retained in the graph and its use during inference.

Our main contributions are summarized as follows:
\begin{itemize}
\item We distinguish the information preserved in a human-centric video graph from a reasoning model's ability to use that information, providing a clearer basis for evaluating graph-supervised video reasoning.

\item We introduce a traceable video representation that combines question-guided SAM~3 participant tracks with temporally localized annotations of attributes, actions, emotions, interactions, and intentions.

\item We develop a positive-only annotation protocol in which Qwen3-VL proposes semantic claims, HumanOmniV2 checks their support against video and available audio, and human quality control reviews identity and temporal grounding.

\item THG-Omni outperforms the evaluated baselines across all three IntentBench tasks, achieving strong results in social reasoning, emotion understanding, and deception perception. Additional graph-input experiments confirm that the proposed representation retains answer-relevant evidence and that graph-specific training helps the model use it.
\end{itemize}

\section{Related Work}
\paragraph{Human-centric Video Understanding.}
Human-centric video understanding includes social reasoning, emotion recognition, intention inference, and deception-related judgments \cite{zadeh2019social,cai2025mdpe}. 
Action-intention relationships have also been studied through language-guided localization of unintentional actions~\cite{xu2025unintentional}.
Answers may depend on facial expressions, motion, speech, vocal delivery, and interactions among participants. Because these cues can occur at different times and across modalities, recent models combine visual, acoustic, and textual evidence \cite{cheng2024emotion,jin2024chatunivi,li2025graphsmile}.  Long-video studies further highlight the difficulty of retaining and retrieving relevant evidence across extended contexts \cite{wu2024longvideobench,fu2025video,
hannan2025revisionllm,moviechatplus2026}. HumanOmniV2's IntentBench evaluates such reasoning through Social-IQ, EMER, and MDPE, which have different prediction formats and evidence requirements \cite{yang2025humanomniv2}. These tasks motivate representations that preserve participant identity and the time at which each observation holds.

\paragraph{Video Grounding and Scene Graphs.}
Promptable segmentation provides spatial grounding: Segment Anything introduced image-level prompting, SAM~2 propagated masks through video, and SAM~3 added concept-prompted detection and tracking \cite{kirillov2023segment,ravi2025sam2,carion2026sam}. Persistent identity across shots remains a separate challenge from mask localization \cite{gao2025multiple}.  Recent surveys have also summarized temporal grounding methods based on multimodal large language models~\cite{wu2026vtgsurvey}.
Video scene graphs organize entities and changing relations over time \cite{huang2024vtimellm}. Existing approaches use temporal caption alignment or neuro-symbolic learning to obtain such structure \cite{kim2025weakly,huang2025laser}. Recent studies further investigate dynamic video scene graph generation and open-vocabulary visual relationship detection \cite{chen2025diffvsgg,wang2025openvocabulary}.
However, a well-formed graph can still assign an action or attribute to the wrong person. Our representation therefore links claims to tracked participants and temporal evidence, and evaluates semantic support separately from structural validity.

\paragraph{Structured Reasoning and Annotation Verification.}
Multimodal knowledge graphs and structure-aware vision-language models expose entities and relations for reasoning \cite{lee2024multimodal,huang2024structure}.  Video-language methods also use localized supervision or question-conditioned evidence acquisition to improve temporal grounding
\cite{zeng2025timesuite,wang2024videoagent, deng2025seq2time,chen2025causal,li2024visualprompt}.
Unlike knowledge-graph completion, which predicts missing relations from entity representations \cite{zhang2024native}, our task uses observations grounded in video trajectories and trains an omni-modal model through graph-specific supervision with low-rank adaptation \cite{hu2021lora,wang2025reasoning}. 
Visually grounded video question answering examines
whether model predictions are supported by relevant
video evidence~\cite{xiao2024grounded}. For semantic fields, Qwen3-VL generates candidate claims and HumanOmniV2 checks their support against the source video and available audio. Prior work on multimodal critics and model-generated feedback motivates this verification step, but model agreement does not guarantee correctness \cite{xiong2025lllava,yu2025rlaif}. Human review therefore remains necessary for identity binding, temporal extent, and interaction arguments.

\paragraph{Graph Construction and Graph Utilization.}
Existing video grounding methods provide spatial and temporal evidence, while video scene graphs organize participants and their relations. However, these two components do not directly address how an omni-modal reasoning model uses the resulting graph. A graph may contain correct participant attributes, actions, and interactions, but the reasoning model may not associate these fields with the question. In addition, a semantic claim may be assigned to the wrong participant or temporal interval even when the graph satisfies its structural constraints.

Prior studies have explored structured video
representations for video-language understanding.
Finsta~\cite{fei2024finsta} improves video-language
alignment through fine-grained scene graph structures,
while CoVGT~\cite{xiao2023covgt} uses a dynamic graph
transformer to capture object relations for video
question answering. More recent graph-based methods,
including STEP~\cite{qiu2025step} and
HyperGLM~\cite{nguyen2025hyperglm}, further explore
structured representations for video reasoning.
Unlike these approaches, THG-Omni focuses on
identity-bound semantic claims, claim verification,
and graph-specific supervision for an omni-modal
reasoning model. Compositional video question answering has also been studied through question decomposition and video evidence alignment~\cite{liao2024align}. Video-grounded entailment tree reasoning provides another approach to organizing intermediate evidence for video question answering
\cite{liu2025entailment}. These limitations indicate that graph construction and graph utilization require different forms of supervision. Graph construction requires participant identity, temporal grounding, and claim verification. Graph utilization requires the reasoning model to interpret graph fields and associate them with the source video and available audio. THG-Omni addresses these requirements by combining SAM3 person tracks, Qwen3-VL semantic annotations, and graph-specific supervision. The framework also separates the information retained in the predicted graph from the effect of graph input on video reasoning. This distinction is examined through the MER24 diagnostic and the graph-input comparisons on IntentBench and NExT-QA.

\section{Method}
\label{method}
\subsection{Overview}
Our framework constructs a human-centric graph from video evidence and uses the graph to supervise omni-modal video reasoning. SAM3 grounds and tracks people to obtain temporal and spatial evidence. Qwen3-VL then describes visible attributes, actions, affective states, interactions, and possible intentions using grounded person identifiers. The two sources are assembled into a graph while retaining the provenance of each claim.
Human quality control and HumanOmniV2 verification filter semantic claims before training. The accepted claims, together with deterministic questions derived from person tracks, form graph-specific supervision. At evaluation time, the model receives a predicted graph from the evaluation video; these evaluation annotations are not manually corrected or claim-verified. Figure~\ref{fig:framework} summarizes these stages.

\subsection{Question-Guided Human Grounding}
We first parse the question to identify explicit human references, grammatical number, and the requested semantic category. The resulting prompt set may include \textit{person}, \textit{man}, \textit{woman}, \textit{boy}, and \textit{girl}. Answer candidates are not used to construct the grounding prompts. When several questions share the same video and prompt set, the corresponding grounding record is computed once and reused.
Given a video $V$ and a question $Q$, geometric grounding produces
\begin{equation}
T=g_{\mathrm{SAM3}}\bigl(V,p(Q)\bigr),
\end{equation}
where $p(Q)$ denotes the human references extracted from the question, and $T$ contains person tracks and geometric measurements.

SAM3 performs promptable person segmentation and temporal mask propagation \cite{carion2026sam}. Each video is sampled at up to 20 frames for construction of the geometric graph. Candidate masks are filtered using fixed constraints on mask area, bounding-box area, side length, and aspect ratio. Intersection-over-union and containment tests remove redundant detections. Track association uses mask overlap, normalized center displacement, and a fixed maximum temporal gap.

For track $i$ at sampled frame $t$, the observation is:
\begin{equation}
o_i^t=(t,b_i^t,c_i^t,a_i^t),
\end{equation}
where $b_i^t$ is the bounding box, $c_i^t$ is its center, and $a_i^t$ is the mask-area ratio relative to the frame. The trajectory:
\begin{equation}
\mathcal{T}_i=\{o_i^t:t\in F_i\}
\end{equation}
records the visible extent and spatial motion of the participant. At this stage, the system does not assign emotions, intentions, or social roles.

\subsection{Deterministic Spatio-Temporal Graph}
The geometric component is represented as:
\begin{equation}
G_T=(\mathcal{V}_T,\mathcal{E}_T).
\end{equation}
Each valid human track defines a node containing its prompt category, temporal extent, visibility coverage, and compact trajectory statistics. Pairwise relations are calculated only when two tracks are visible in a sufficient number of common sampled frames.

For tracks $i$ and $j$, their center distance at frame $t$ is normalized by the frame diagonal:
\begin{equation}
d_{ij}^{t}
=
\frac{\lVert c_i^t-c_j^t\rVert_2}
{\sqrt{W^2+H^2}},
\end{equation}
where $W$ and $H$ are the frame width and height. The graph records co-visibility, relative horizontal and vertical position, mask overlap, normalized distance, and distance change. A video-level relation is retained only when its temporal support exceeds a fixed threshold. The supporting frames and source tracks remain attached to each relation, allowing every deterministic graph field to be traced back to the video.

The deterministic graph records geometric evidence without assigning semantic labels. In particular, co-visibility indicates whether two person tracks are observed in the same sampled frames, while normalized distance describes their relative spatial separation. Distance change provides additional information about how the separation varies over time. These fields are computed from person tracks and do not depend on the semantic annotations generated by Qwen3-VL.

Temporal support is important when constructing pairwise relations. Two participants may appear close to each other in a small number of frames without remaining close throughout the video. Similarly, a participant may disappear because of occlusion or a camera transition. The graph therefore retains the sampled frames supporting each relation instead of treating every observed spatial relation as valid for the complete video. This allows the temporal extent of a relation to be checked against the corresponding person tracks. 
Related work on spatio-temporal video grounding has also investigated supervision from sparsely annotated frames~\cite{liu2025singleframe}.

The geometric fields also have clear limitations. A small distance between two participants does not establish that they are interacting, and a change in distance does not directly indicate an intention or emotional state. Such information requires semantic annotations supported by video evidence. THG-Omni therefore keeps geometric evidence and semantic claims as separate components before graph assembly. This separation allows deterministic graph questions to use track statistics while semantic questions use claims that pass the verification process.

\subsection{Whole-Video Semantic Annotation and Graph Assembly}
Geometric tracks do not identify actions, affect, or interactions. Qwen3-VL-8B therefore annotates the complete video and an automatically generated contact sheet, using grounded SAM3 identifiers as identity anchors. Let $H(T)$ be the grounded identifier set and $L_C$ a task-specific candidate vocabulary, when one exists. The semantic annotations are generated as
\begin{equation}
S=g_{\mathrm{ann}}\bigl(V,\operatorname{contact}(V),H(T),L_C\bigr).
\end{equation}
The annotator receives neither the reference answer nor a graph corrected using that answer. For MER24, $L_C$ contains the six candidate emotion labels associated with the video; Social-IQ uses free-form descriptions because it has no corresponding closed inventory.

We represent a semantic claim as
\begin{equation}
\hspace{-0.2cm}s_k=(u_k,\ell_k,I_k,E_k,\gamma_k),\qquad\\
I_k=[r_k^{\mathrm{s}},r_k^{\mathrm{e}}],
0\leq r_k^{\mathrm{s}}\leq r_k^{\mathrm{e}}\leq 1,
\end{equation}
where $u_k$ is a person identifier or an ordered pair of people for an interaction, $\ell_k$ is the claimed attribute, affect, action, interaction, or intent, $I_k$ is its normalized video interval, $E_k$ contains one to three supporting evidence ratios in $[0,1]$, and $\gamma_k$ is a confidence category. The schema also records the people present and annotation limits. Every argument of a claim must name a declared person, and self-interactions are excluded. Appearance claims crossing a camera transition require renewed visual evidence.

An identity map $\pi$ attaches an annotation identifier to its SAM3 track whenever the track is available. The annotator may introduce a neutral identifier for an additional visible person, but the resulting node remains semantic-only until geometric support exists. Invalid or ambiguous bindings are excluded rather than assigned to a convenient track. The assembled graph is
\begin{equation}
\begin{aligned}
G&=g_{\mathrm{merge}}(G_T,S;\pi)\\
 &=\bigl(\mathcal V_T\cup\mathcal V_S,\,
         \mathcal E_T\cup\mathcal E_S\bigr),
\end{aligned}
\end{equation}
where geometric fields in $(\mathcal V_T,\mathcal E_T)$ retain their track and frame provenance, while semantic fields in $(\mathcal V_S,\mathcal E_S)$ retain their claim, interval, evidence ratios, and confidence. A semantic-only person has no deterministic geometric fields. Whole-video annotation is used in the formal pipeline; a fixed-window pilot produced fragmented identities when participants overlapped.

The identity map $\pi$ connects the person identifiers used by Qwen3-VL with the person tracks obtained from SAM3. This step is necessary because geometric grounding and semantic annotation provide different information about the same video. SAM3 identifies visible participants and their spatial positions, while Qwen3-VL describes participant attributes, actions, affective states, interactions, and possible intentions. Without identity binding, a semantic claim may describe the correct event but refer to the wrong participant.

Identity binding is particularly important when several participants appear in the same video. A semantic claim describing an interaction must identify the participants involved, and its temporal interval must correspond to the period in which the interaction is supported. When the available person tracks do not provide a valid identity binding, the annotation is not attached to an unrelated track. A semantic-only person may be retained, but deterministic geometric fields are not assigned without geometric support.

The assembled graph preserves the source information of both components. Geometric fields retain their person tracks and supporting frames, while semantic claims retain their person identifiers, temporal intervals, evidence ratios, and confidence categories. These fields allow a graph question to refer to a specific participant or event and retain the evidence associated with that question. The resulting human-centric video graph is therefore used not only as additional input to the reasoning model, but also as a source of graph-specific supervision.

\subsection{Training-Only Quality Control and Claim Verification}
Human inspection of development outputs checks person binding, temporal extent, label validity, and visual support before large-scale construction. Schema normalization is conservative: it may standardize a supported format, set an inapplicable optional label to \texttt{null}, or retain existing evidence points from an overlong list. It does not create claims. Malformed person bindings, invalid intervals, non-human nodes, and unsupported interaction arguments are excluded; changes and exclusions are logged. Appearance changes across camera transitions receive interval-level review.

HumanOmniV2 independently checks each remaining semantic claim against the source video and available audio \cite{yang2025humanomniv2}. It sees neither the benchmark answer nor a synthetic negation. For a candidate $s_k$, define
\begin{equation}
\begin{aligned}
z_k &=
\operatorname{Verify}_{\mathrm{HumanOmniV2}}(s_k,V,A),\\
z_k &\in
\{\texttt{supported},\texttt{unsupported},\texttt{uncertain}\}.
\end{aligned}
\end{equation}
The retained semantic collection is
\begin{equation}
\begin{aligned}
S^{+}=\bigl\{s_k\in S:\;&
\operatorname{schema\_valid}(s_k)\\
&\hspace{-1.2cm}\land\operatorname{identity\_valid}(s_k)
\land z_k=\texttt{supported}\bigr\}.
\end{aligned}
\end{equation}

Manually rejected, unsupported, uncertain, and malformed claims are excluded. When a multi-claim response cannot be parsed, only its affected verification unit is retried; original outputs and parser repairs are preserved. This positive-only protocol avoids the lexical shortcuts and equivalent counterfactuals observed in an earlier original-versus-negated discrimination setting. Both human review and claim verification apply to training and development construction, not evaluation graphs.

Schema validation, identity checks, and claim verification address different sources of errors. Schema validation determines whether a semantic claim follows the required format, including valid person identifiers, temporal intervals, and interaction arguments. However, a schema-valid claim may still describe an event that is not supported by the video. Identity checks further examine whether the claim refers to a declared participant and whether the available person tracks support the identity binding. These checks reduce invalid graph construction but do not establish the correctness of every semantic claim.

Claim verification therefore provides an additional step. HumanOmniV2 evaluates each candidate semantic claim using the source video and available audio and assigns a supported, unsupported, or uncertain verdict. Only supported claims that also satisfy schema and identity requirements are retained for semantic question construction. Unsupported and uncertain claims are excluded rather than converted into negative training examples. This is consistent with the positive-only annotation protocol used in THG-Omni.

The verification results should be interpreted as model-based assessments rather than human-confirmed labels. A supported verdict indicates that the verifier found evidence for the candidate claim, but it does not guarantee that the claim is correct. Human quality control is therefore used to inspect identity binding, temporal extent, and visual support during training-data construction. These checks are not applied as manual corrections to evaluation graphs. During inference, the reasoning model receives the predicted graph constructed from the evaluation video without verifier verdicts.

\subsection{Graph-Specific Supervision}
The final training collection combines deterministic graph questions with verified semantic questions. Deterministic targets are computed from frozen track statistics and cover temporal duration, first appearance, relative presence, co-visibility, overlap, distance change, and directional relations. Semantic targets are derived only from accepted positive claims and cover visible attributes, actions, emotions, interactions, and intentions.

For video $V$, optional audio $A$, question $Q$, answer candidates $C$, and graph $G$, the downstream prediction is
\begin{equation}
\hat{Y}=f_{\theta,\phi}(V,A,Q,C,G),
\end{equation}
where $\theta$ denotes the frozen omni-modal backbone and $\phi$ denotes the trainable LoRA parameters.

Training first fits an answer-supervised LoRA adapter (Task-LoRA), merges it into the frozen backbone, and then performs one epoch of joint adaptation. The joint collection includes answer questions with and without graph input, verified graph questions, and matched or mismatched graph identification. With the training examples denoted by $\mathcal{D}_{\mathrm{joint}}$, the joint-stage objective is
\begin{equation}
\mathcal{L}_{\mathrm{joint}}
=-\sum_{(V,A,G,Q,Y)\in\mathcal{D}_{\mathrm{joint}}}
\log P_{\theta,\phi}\bigl(Y\mid V,A,G,Q\bigr),
\end{equation}
where $G$ may be omitted for graph-dropout examples. For multiple-choice questions, the candidate set $C$ is included in $Q$ in the training objective. 

LoRA modules are attached to the language-model attention projections \cite{hu2021lora}. The visual tower, audio tower, and base language-model parameters remain frozen. Human review and cross-model verification are restricted to training and development data construction. During evaluation, the model receives only a graph predicted from the evaluation video. No evaluation annotation is manually corrected, and no verifier verdict is used as an inference input.

The two training stages have different purposes. Task-LoRA first adapts Qwen2.5-Omni-7B to the answer format of the downstream tasks. The resulting adapter is merged into the backbone before joint training. This provides a task-adapted checkpoint for the subsequent graph-specific supervision. The joint stage then trains the model using task questions and graph questions rather than replacing the original task supervision with graph questions alone.

The deterministic graph questions and verified semantic questions also provide different supervision. Deterministic questions are constructed from person tracks and geometric relations. Their answers are computed from the graph fields, including temporal duration, visibility, co-visibility, and distance change. These questions teach the model to interpret geometric information that can be directly calculated from the person tracks. Verified semantic questions are constructed from accepted semantic claims and cover visible attributes, actions, emotions, interactions, and intentions. These questions teach the model to associate semantic graph fields with the corresponding participants and temporal intervals.

The joint collection further includes task questions with graph dropout and task questions with matched graphs. Graph-dropout examples require the model to answer using the source video and available audio without graph input. Matched-graph examples provide the predicted graph as additional input for the same type of video reasoning task. Including both conditions allows the model to encounter questions with and without graph input during training. This is important because a predicted graph may not contain all the information required to answer a question.

Matched or mismatched graph-identification examples provide supervision about graph--video correspondence. A matched graph is constructed from the corresponding video, while a mismatched graph contains information from another video. These examples require the model to distinguish graph input associated with the source video from graph input that does not correspond to it. The joint training objective combines these supervision types using the same conditional language-modeling loss. The different training examples therefore contribute to graph interpretation, task question answering, and graph--video correspondence within one training stage.

\begin{figure}[!t]
\centering
\begin{minipage}{0.98\columnwidth}
\footnotesize
\hrule
\vspace{0.5ex}
\noindent\textbf{Algorithm 1: Traceable Human-Centric Graph Supervision}
\vspace{0.5ex}
\hrule
\vspace{0.5ex}

\noindent\textbf{Require:}
Training examples $\{(V,A,Q,Y)\}$, frozen backbone
$f_{\theta}$, and candidate vocabulary $L_C$ when available.

\noindent\textbf{Ensure:}
Task-adapted backbone and joint-stage LoRA parameters $\phi$.

\vspace{0.4ex}
\setlength{\tabcolsep}{2pt}
\renewcommand{\arraystretch}{1.05}
\begin{tabular}{@{}p{0.065\linewidth}@{\hspace{0.5em}}p{0.89\linewidth}@{}}

1: &
Initialize $\mathcal{D}_{\mathrm{graph}}\gets\emptyset$. \\

2: &
\textbf{for} each training video $V$ and grounding question $Q$
\textbf{do} \\

3: &
\hspace{0.7em} Obtain tracks
$T\gets g_{\mathrm{SAM3}}(V,p(Q))$. \\

4: &
\hspace{0.7em} Construct
$G_T=(\mathcal{V}_T,\mathcal{E}_T)$
from track statistics and spatial relations. \\

5: &
\hspace{0.7em} Generate semantic claims: \\
& \hspace{0.7em}
$\displaystyle S\gets g_{\mathrm{ann}}
(V,\operatorname{contact}(V),H(T),L_C)$. \\

6: &
\hspace{0.7em} Review identity binding, temporal extent,
label validity, and visual support. \\

7: &
\hspace{0.7em} \textbf{for} each remaining claim $s_k\in S$
\textbf{do} \\

8: &
\hspace{1.4em} Compute verifier result: \\
& \hspace{1.4em}
$z_k\gets\operatorname{Verify}_{\mathrm{HumanOmniV2}}
(s_k,V,A)$. \\

9: &
\hspace{0.7em} \textbf{end for} \\

10: &
\hspace{0.7em} Retain supported claims $S^+$
according to (10). \\

11: &
\hspace{0.7em} Assemble
$G\gets g_{\mathrm{merge}}(G_T,S^+;\pi)$. \\

12: &
\hspace{0.7em} Add deterministic questions from $G_T$
and verified semantic questions from $S^+$
to $\mathcal{D}_{\mathrm{graph}}$. \\

13: &
\textbf{end for} \\

14: &
Train a task question-answering LoRA adapter
and merge it into the backbone. \\

15: &
Construct $\mathcal{D}_{\mathrm{joint}}$ from task questions
with graph dropout or matched graphs, verified graph
questions, and matched or mismatched graph-identification
examples. \\

16: &
Optimize $\phi$ for one epoch with $\theta$ frozen
using the joint objective in (12). \\

17: &
\textbf{return} the task-adapted backbone
and joint-stage adapter $\phi$. \\

\end{tabular}

\vspace{0.5ex}
\hrule
\end{minipage}
\end{figure}

\noindent
For graph-dropout examples, $G$ is omitted.
Answer candidates $C$ are included in $Q$.
Human review and claim verification are restricted
to training and development construction.

\section{Experiments}\label{sec:experiments}
\subsection{Experimental Setup}

We evaluate THG-Omni on human-centric video reasoning tasks. MER24 contains 150 multi-label emotion samples and is used to test the information retained in the scene graph. IntentBench covers social reasoning, emotion understanding, and deception perception. Its original manifest contains 2,689 questions from 633 videos. One MDPE video is excluded by a stable task identifier because it repeatedly exhausts GPU memory even when processed alone on an NVIDIA L20. The resulting evaluation set contains 2,688 aligned question--graph pairs from 632 videos: 2,356 Social-IQ questions, 133 EMER questions, and 199 MDPE questions. The exclusion is fixed before prediction and scoring. NExT-QA~\cite{xiao2021next} is additionally used to evaluate video question answering across causal, temporal, and descriptive reasoning tasks. The test set contains 8,564 questions from 1,000 videos, and the validation set contains 4,996 questions from 570 videos. Each question has five answer options. Unlike IntentBench, which includes both single-label and multi-label tasks, NExT-QA is evaluated using multiple-choice accuracy.

Graph-specific supervision is constructed from person tracks and accepted semantic claims. The graph-question collection contains 7,339 training examples, comprising 1,555 deterministic temporal and spatial questions and 5,784 verified positive semantic questions; its development collection contains 1,146 examples. The final joint-training stage presents 2,887 examples in one epoch, combining answer questions, graph questions, and matched or mismatched graph examples. The graph-question training and development partitions do not share videos. 
Qwen2.5-Omni-7B serves as the reasoning backbone. SAM3 provides person tracks and geometric relations, Qwen3-VL-8B generates whole-video semantic annotations, and HumanOmniV2 assesses the support for candidate semantic claims. In the video reasoning pipeline, frames are sampled at 0.25 FPS, generation is deterministic, and available audio is provided to the reasoning model. The annotation model receives video frames and a contact sheet without audio. The verifier may use audio when assessing a claim. LoRA modules are applied to the language attention projections; the visual tower, audio tower, and base language-model weights remain frozen. Additional training and checkpoint details are given in Section~\ref{sec:adapter_controls}.

Social reasoning and deception perception are measured by accuracy. Emotion understanding is measured by sample-level multi-label F1. F1 is computed from each question's predicted and reference label sets and then averaged over questions. Exact set match requires equality of the two sets. For MER24, Table~\ref{tab:mer_graph_content} reports this sample-averaged F1 as macro F1; micro F1 instead pools true positives, false positives, and false negatives across all samples before computing F1. Its set precision and recall are averaged separately over samples. The overall IntentBench score is a sample-weighted aggregation across the 2,688 questions. Every condition in an IntentBench comparison uses the same questions and evaluation videos.

\subsection{Main Experiments}

\begin{figure*}[t]
    \centering
\includegraphics[width=0.98\textwidth]{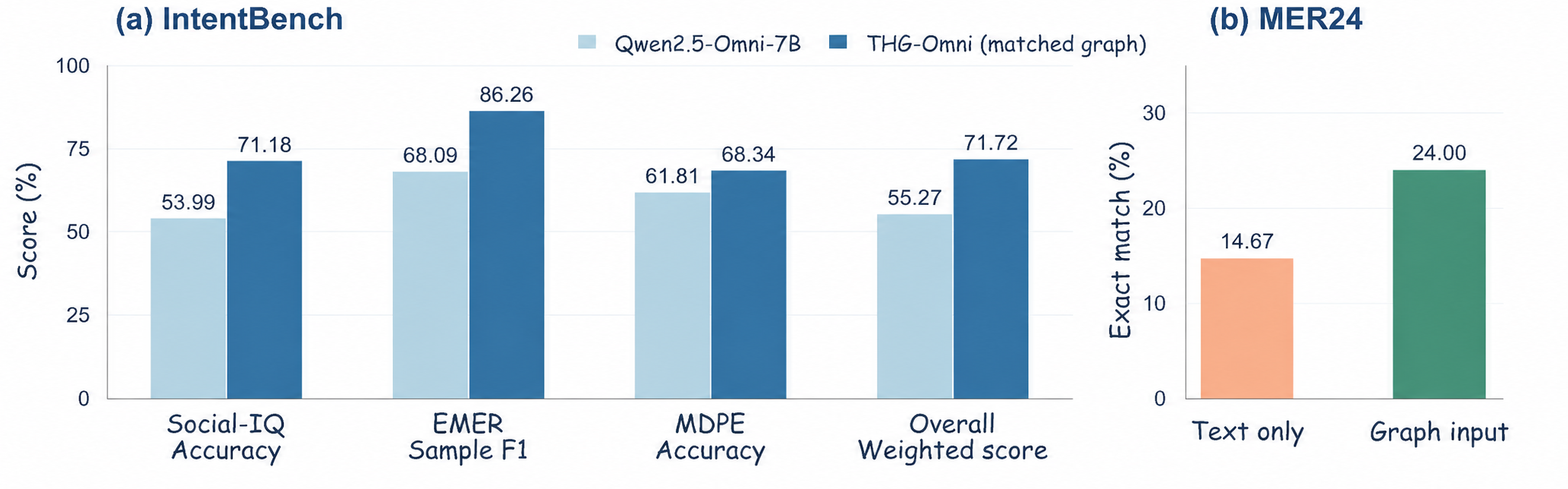}
    \caption{IntentBench results and MER24 graph diagnostic. (a) Qwen2.5-Omni-7B and video-matched THG-Omni are compared on 2,688 questions. (b) On 150 MER24 samples without source video or audio, graph input improves exact set match over text-only input.}
    \label{fig:results}
\end{figure*}

Table~\ref{tab:thg_main} reports IntentBench results on the same 2,688 questions. The matched THG-Omni achieves 71.18\% Social-IQ accuracy and 68.34\% MDPE accuracy, compared with 53.99\% and 61.81\% for Qwen2.5-Omni-7B. EMER sample F1 is 86.26\% versus 68.09\% for Qwen2.5-Omni-7B, while EMER exact match is 66.17\% versus 41.35\%. Our method has a sample-weighted overall score of 71.72\%. Figure~\ref{fig:results} visualizes the baseline and THG-Omni scores alongside the separate MER24 graph-information diagnostic. Graph-SFT denotes a separate LoRA adapter trained for graph-field question answering and evaluated with either no graph or a compact graph; these two rows share weights. THG-Omni denotes the staged, joint-training model described in Section~\ref{method}. The THG-Omni rows compare graph-input conditions of the same trained model, whereas Graph-SFT uses
a separate adapter.

LLaVA-NeXT-Video-7B-32K provides an additional video-only baseline on the same 2,688 questions. It receives eight uniformly sampled frames and the question with its answer options, without audio or graph input, and obtains 31.58\% Social-IQ accuracy, 29.68\% EMER sample F1, and 39.20\% MDPE accuracy. These scores use less input evidence than the audio-video runs and should be interpreted with that modality difference in mind. Responses with a recoverable answer option are scored, while unparseable responses are counted as incorrect.

HumanOmniV2~\cite{yang2025humanomniv2} is also evaluated without graph input on the same 2,688 questions. It obtains 51.57\% Social-IQ accuracy, 80.03\% EMER sample F1, and 53.77\% MDPE accuracy. THG-Omni improves EMER sample F1 to 86.26\% on the same questions.

\begin{table*}[t]
\centering
\caption{IntentBench results on 2,688 questions. Overall is the sample-weighted aggregate of Social-IQ accuracy, EMER sample F1, and MDPE accuracy. LLaVA-NeXT-Video uses video without audio. Graph-SFT uses a distinct adapter; the THG-Omni rows report different graph-input conditions for the same trained model.}
\label{tab:thg_main}
\footnotesize
\setlength{\tabcolsep}{2pt}
\begin{tabularx}{\linewidth}{@{}>{\raggedright\arraybackslash}X*{5}{>{\centering\arraybackslash}p{0.112\linewidth}}@{}}
\toprule
 & Social-IQ & \multicolumn{2}{c}{EMER} & MDPE & Overall \\
\cmidrule(lr){3-4}
Model / input & Accuracy & Sample F1 & Exact & Accuracy & Score \\
\midrule
LLaVA-NeXT-Video ~\cite{zhang2024llava}
    & 31.58 & 29.68 & 12.03 & 39.20 & 32.05 \\
HumanOmniV2~\cite{yang2025humanomniv2}
    & 51.57 & 80.03 & 54.89 & 53.77 & 53.14 \\
Qwen2.5-Omni-7B~\cite{xu2025qwen25omnitechnicalreport}
    & 53.99 & 68.09 & 41.35 & 61.81 & 55.27 \\
\midrule
Graph-SFT (w/o graph)
    & 49.66 & 70.52 & 40.60 & 60.30 & 51.48 \\
Graph-SFT (compact graph)
    & 56.66 & 71.32 & 46.62 & 61.31 & 57.73 \\
\midrule
THG-Omni (w/o graph)
    & 67.28 & 84.63 & 66.17 & 62.31 & 67.77 \\
THG-Omni (shuffled graph)
    & 67.11 & 81.29 & 61.65 & 61.31 & 67.38 \\
\rowcolor{gray!8}
\textbf{THG-Omni (matched graph; Ours)}
    & \textbf{71.18} & \textbf{86.26} & \textbf{66.17}
    & \textbf{68.34} & \textbf{71.72} \\
\bottomrule
\end{tabularx}
\end{table*}

\noindent\textbf{Social-IQ question-type analysis.}
Social-IQ contains six question groups: Why, How, What, When, Who/Which, and Other. Table~\ref{tab:socialiq_types} reports accuracy across these question types.

\begin{table*}[t]
\centering
\caption{Social-IQ accuracy (\%) across six question types on 2,356 questions.}
\label{tab:socialiq_types}
\footnotesize
\setlength{\tabcolsep}{2pt}
\begin{tabularx}{\linewidth}{@{}>{\raggedright\arraybackslash}p{0.26\linewidth}*{7}{>{\centering\arraybackslash}X}@{}}
\toprule
Method & \shortstack{Why\\(698)} & \shortstack{How\\(632)} & \shortstack{What\\(480)} & \shortstack{When\\(14)} & \shortstack{Who/Which\\(25)} & \shortstack{Other\\(507)} & Overall \\
\midrule
Qwen2.5-Omni-7B~\cite{xu2025qwen25omnitechnicalreport}
& 53.44 & 54.11 & 53.54 & 57.14 & 64.00 & 54.44 & 53.99 \\

Graph-SFT (w/o graph)
& 47.99 & 49.05 & 50.00 & 50.00 & 56.00 & 52.07 & 49.66 \\

Graph-SFT (compact graph)
& 55.01 & 56.01 & 57.08 & 57.14 & 64.00 & 58.97 & 56.66 \\

THG-Omni (w/o graph)
& 66.19 & 65.66 & 66.88 & 64.29 & 72.00 & 71.01 & 67.28 \\

THG-Omni (shuffled graph)
& 66.48 & 65.19 & 65.83 & 64.29 & 72.00 & 71.40 & 67.11 \\

\rowcolor{gray!8}
\textbf{THG-Omni (matched graph)}
& \textbf{69.63} & \textbf{69.78} & \textbf{70.83}
& \textbf{71.43} & \textbf{80.00} & \textbf{74.95}
& \textbf{71.18} \\
\bottomrule
\end{tabularx}
\end{table*}

\begin{figure*}[t]
    \centering
    \includegraphics[width=0.98\textwidth]{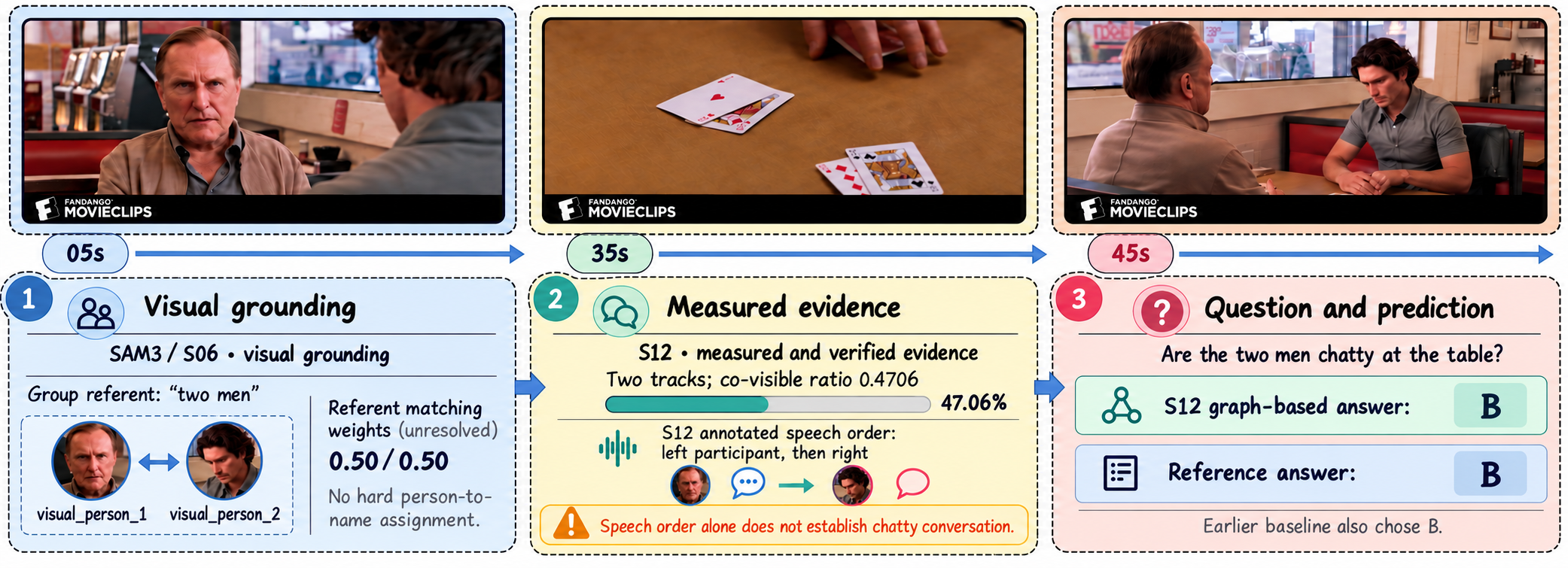}
    \caption{Illustrative Social-IQ video-to-graph example.
    The video frames show two participants and their
    interaction during a card game. SAM3 provides person
    tracks and geometric evidence, while semantic
    annotations describe the observed interaction.
    The example illustrates participant grounding and
    temporally localized visual evidence associated
    with the Social-IQ question. The reference answer is B.}
    \label{fig:social_iq_case}
\end{figure*}

\noindent\textbf{EMER multi-label metrics.}
Table~\ref{tab:emer_multilabel} reports precision, recall, sample-level F1, and exact-match performance.

\begin{table}[t]
\centering
\caption{EMER multi-label performance on 133 examples (\%).}
\label{tab:emer_multilabel}
\footnotesize
\setlength{\tabcolsep}{3pt}
\begin{tabularx}{\linewidth}{@{}>{\raggedright\arraybackslash}p{0.40\linewidth}*{4}{>{\centering\arraybackslash}X}@{}}
\toprule
Method & Precision & Recall & Sample F1 & Exact \\
\midrule
Qwen2.5-Omni-7B~\cite{xu2025qwen25omnitechnicalreport}
& 74.00 & 68.53 & 68.09 & 41.35 \\

Graph-SFT (w/o graph)
& 76.20 & 70.10 & 70.52 & 40.60 \\

Graph-SFT (compact graph)
& 78.10 & 70.80 & 71.32 & 46.62 \\

THG-Omni (w/o graph)
& 89.54 & 83.97 & 84.63 & 66.17 \\

THG-Omni (shuffled graph)
& 87.97 & 79.52 & 81.29 & 61.65 \\

\rowcolor{gray!8}
\textbf{THG-Omni (matched graph)}
& \textbf{93.36} & 83.91 & \textbf{86.26} & 66.17 \\
\bottomrule
\end{tabularx}
\end{table}

\begin{figure*}[t]
\centering
\includegraphics[width=\textwidth]{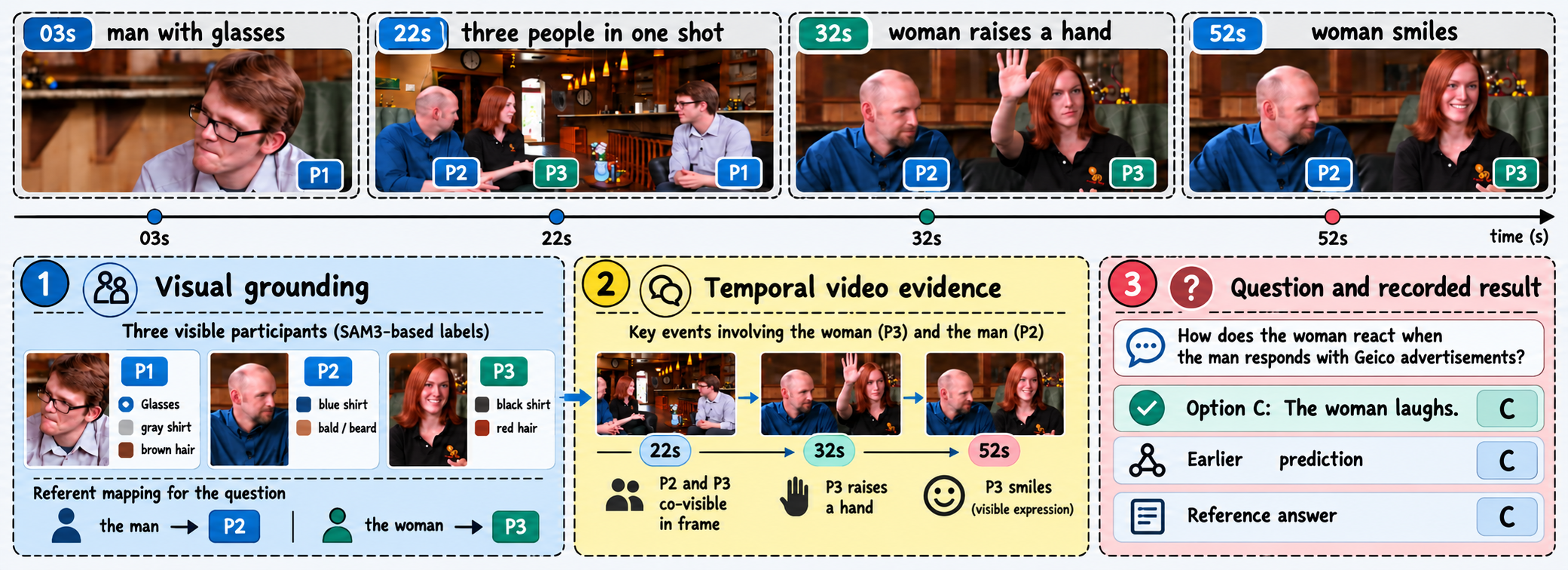}
\caption{Illustrative Social-IQ video-to-graph example.
The video frames show the visible participants and
the woman's later reaction. The example illustrates
participant grounding and temporally localized visual
evidence associated with the Social-IQ question.
The reference answer is C. This example is used to
illustrate evidence organization rather than a
performance improvement.}
\label{fig:app_social_iq_example}
\end{figure*}

\noindent\textbf{Graph-input sensitivity on IntentBench.}
Rather than repeating the absolute scores in Table~\ref{tab:thg_main},
Table~\ref{tab:intentbench_graph_delta} reports the differences between
specified conditions. For the fixed Graph-SFT adapter, adding its compact
graph improves Social-IQ accuracy by 7.00 percentage points and EMER exact
match by 6.02 points. For THG-Omni, the reported matched-graph condition
exceeds the no-graph condition by 3.90 points on Social-IQ and 6.03 points
on MDPE. The matched and shuffled THG-Omni results were reported from
different evaluation runs; these differences should not be interpreted as
paired statistical estimates. EMER differences refer to sample-level F1,
not classification accuracy.

\begin{figure*}[t]
\centering
\includegraphics[width=\textwidth]{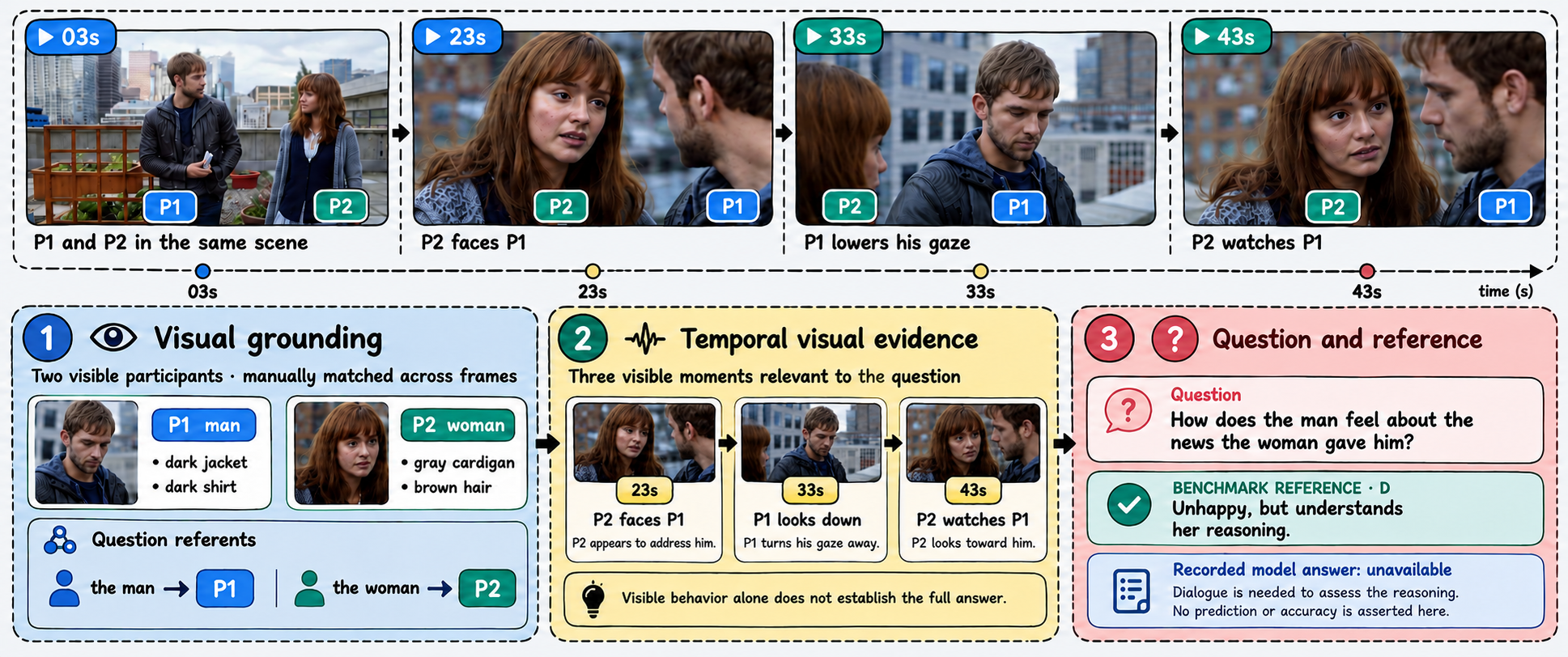}
\caption{Illustrative Social-IQ video-to-graph example.
The video frames show participant grounding and
temporally localized visual evidence relevant to the
question. Participant identities are manually matched
across frames for illustration. The reference answer
is D, while the recorded model answer is unavailable.
This example illustrates evidence organization rather
than a verified model prediction.}
\label{fig:app_social_iq_example2}
\end{figure*}

\begin{table*}[t]
\centering
\caption{IntentBench graph-input sensitivity, expressed as differences in
percentage points between reported conditions. EMER uses sample-level F1
and exact set match; other task columns use accuracy.}
\label{tab:intentbench_graph_delta}
\footnotesize
\setlength{\tabcolsep}{2pt}
\renewcommand{\arraystretch}{1.12}
\begin{tabularx}{\linewidth}{@{}>{\raggedright\arraybackslash}p{0.36\linewidth}*{5}{>{\centering\arraybackslash}X}@{}}
\toprule
Comparison (second minus first) & Social-IQ & EMER F1 & EMER Exact & MDPE & Overall \\
\midrule
Graph-SFT: compact $-$ no graph & +7.00 & +0.80 & +6.02 & +1.01 & +6.25 \\
THG-Omni: matched $-$ no graph & +3.90 & +1.63 & +0.00 & +6.03 & +3.95 \\
THG-Omni: matched $-$ shuffled & +4.07 & +4.97 & +4.52 & +7.03 & +4.34 \\
\bottomrule
\end{tabularx}
\end{table*}

Table~\ref{tab:mer_graph_content} reports a MER24 graph-information diagnostic on 150 samples. Both conditions receive the same question and answer options, without source video or audio; only the graph input differs. Adding the predicted graph increases exact match from 14.67\% to 24.00\% and macro F1 from 42.18\% to 57.18\%. There are 21 rescued and 7 harmed examples (exact McNemar $p=0.0125$), and the bootstrap 95\% interval for the exact-match difference is $[2.67\%,16.00\%]$. This diagnostic tests information recoverable from the graph, rather than improvement to video-conditioned reasoning.

\begin{table*}[t]
\centering
\caption{MER24 graph-information diagnostic on 150 samples. Both conditions exclude source video and audio; the only input difference is the predicted scene graph. All values are percentages. Set precision and recall are sample-level averages.}
\label{tab:mer_graph_content}
\footnotesize
\setlength{\tabcolsep}{3pt}
\begin{tabularx}{\linewidth}{@{}>{\raggedright\arraybackslash}p{0.30\linewidth}*{5}{>{\centering\arraybackslash}X}@{}}
\toprule
Input & Exact & Set precision & Set recall & Macro F1 & Micro F1 \\
\midrule
Question and options & 14.67 & 37.90 & 56.67 & 42.18 & 44.23 \\
Question, options, and graph & \textbf{24.00} & \textbf{54.40} & \textbf{69.33} & \textbf{57.18} & \textbf{56.16} \\
\bottomrule
\end{tabularx}
\end{table*}

\begin{figure*}[!t]
    \centering
    \includegraphics[width=\textwidth]{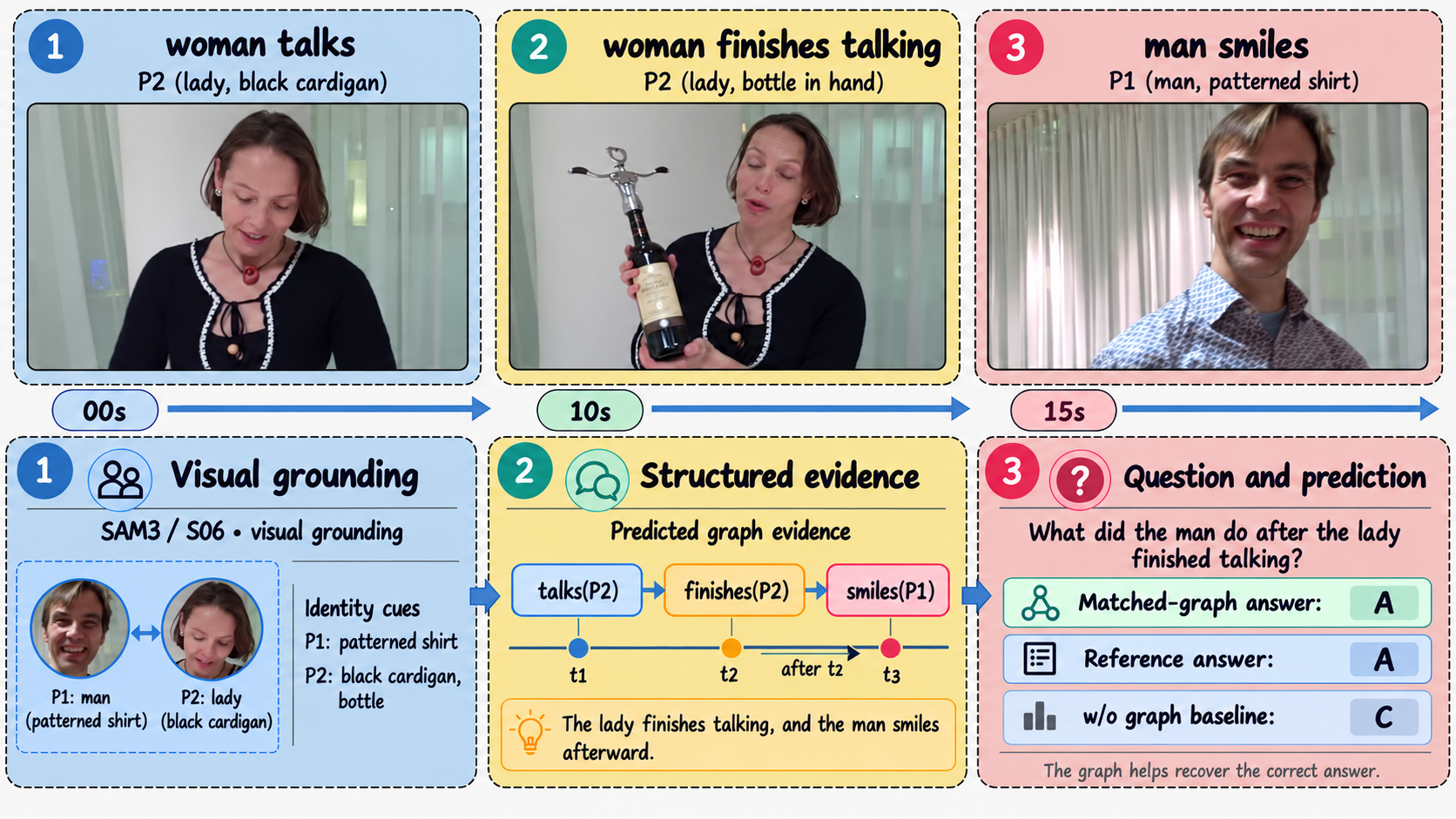}
    \caption{NExT-QA temporal reasoning example from the validation set. Three video frames show the lady talking and the man's subsequent reaction. SAM3 provides participant grounding, while the predicted graph organizes the temporal evidence used for answering the question. With the same Graph-SFT adapter, the no-graph prediction is C, whereas the matched-graph prediction is A, consistent with the reference answer.}
    \label{fig:nextqa_case}
\end{figure*}

\subsection{NExT-QA Evaluation}

We further evaluate video question answering on NExT-QA. The test set contains 8,564 questions from 1,000 videos, and the validation set contains 4,996 questions from 570 videos. SAM3 provides geometric evidence, while Qwen3-VL-8B generates whole-video semantic annotations. One video in each split has an invalid semantic annotation, for which the graph retains only geometric evidence. The resulting predicted graphs cover all 8,564 test questions and 4,996 validation questions. Evaluation graph construction does not use reference answers, question text, answer options, or historical predictions to generate semantic claims.

For the NExT-QA test set, the predicted graphs contain 19,304 visible attribute claims, 17,560 action segments, 6,781 interactions, 10,374 emotion segments, and 9,821 intent hypotheses, with an average graph length of 2,235.68 characters. On the validation set, the corresponding numbers are 11,565, 10,385, 4,188, 5,886, and 5,618, with an average graph length of 2,280.67 characters. These predicted graphs combine SAM3 geometric evidence with Qwen3-VL semantic annotations and are used as additional input to the reasoning model.

Table~\ref{tab:nextqa_results} reports NExT-QA test
accuracy across causal, temporal, and descriptive
questions. Qwen2.5-Omni-7B achieves 77.98\% accuracy,
while Task-LoRA without graph input achieves 76.09\%.
Graph-SFT improves from 77.86\% without graph input
to 82.40\% with matched graph input. THG-Omni achieves
85.10\% without graph input and 89.54\% with matched
graph input. The matched graph improves THG-Omni
accuracy by 4.44 percentage points. Compared with
Qwen2.5-Omni-7B, THG-Omni with matched graph input
improves accuracy by 11.56 percentage points.

These results show that graph-specific supervision
and matched graph input contribute to NExT-QA
performance. THG-Omni without graph input also
outperforms Graph-SFT without graph input, indicating
a difference between the two training conditions.

We further examine graph content on the NExT-QA validation set using the same Graph-SFT adapter. Accuracy increases from 77.58\% without graph input to 78.02\% with the full graph and 78.22\% with the compact graph. The compact graph retains visible attributes, action segments, and interactions, while removing emotion segments and intent hypotheses. Compared with the full graph, the compact graph corrects 66 previously incorrect predictions and changes 56 correct predictions into incorrect ones. Compared with no graph input, it corrects 187 predictions and changes 155 correct predictions into incorrect ones. These results indicate that graph content affects answer prediction, and that removing emotion and intent claims leads to a small improvement on the validation set.

\begin{table*}[t]
\centering
\caption{NExT-QA test accuracy (\%) across reasoning types on 8,564 questions.}
\label{tab:nextqa_results}
\footnotesize
\setlength{\tabcolsep}{2pt}
\renewcommand{\arraystretch}{1.08}
\begin{tabularx}{\linewidth}{
@{}>{\raggedright\arraybackslash}X
*{9}{>{\centering\arraybackslash}p{0.073\linewidth}}@{}
}
\toprule
\multirow{2}{*}{Model / input}
& \multicolumn{2}{c}{Causal}
& \multicolumn{3}{c}{Temporal}
& \multicolumn{3}{c}{Descriptive}
& \multirow{2}{*}{Overall} \\
\cmidrule(lr){2-3}
\cmidrule(lr){4-6}
\cmidrule(lr){7-9}
& CH & CW & TC & TN & TP & DC & DL & DO & \\
\midrule

Qwen2.5-Omni-7B~\cite{xu2025qwen25omnitechnicalreport}
& 76.21 & 79.18 & 77.51 & 72.34 & 75.27
& 61.18 & 93.79 & 85.50 & 77.98 \\

Task-LoRA (no graph)
& 73.74 & 76.99 & 75.28 & 70.19 & 73.12
& 60.87 & 94.00 & 85.17 & 76.09 \\

\midrule

Graph-SFT (no graph)
& 76.39 & 79.18 & 77.25 & 71.77 & 74.19
& 61.18 & 93.58 & 85.67 & 77.86 \\

Graph-SFT (matched graph)
& 81.66 & 83.21 & 82.59 & 76.65 & 80.72
& 70.39 & 95.11 & 88.86 & 82.40 \\

\midrule

THG-Omni (no graph)
& 83.79 & 85.56 & 84.80 & 81.20 & 83.91
& 75.99 & 96.13 & 90.96 & 85.10 \\

\rowcolor{gray!8}
\textbf{THG-Omni (matched graph)}
& \textbf{88.56} & \textbf{89.93} & \textbf{89.13}
& \textbf{86.90} & \textbf{88.02}
& \textbf{82.97} & \textbf{97.41}
& \textbf{93.68} & \textbf{89.54} \\

\bottomrule
\end{tabularx}

\vspace{2pt}
\parbox{\linewidth}{\scriptsize
CH/CW: causal how/why; TC/TN/TP: temporal counting/next/previous;
DC/DL/DO: descriptive counting/location/other.
Category counts: causal 4,502; temporal 2,657; descriptive 1,405.}
\end{table*}

We further examine graph input on the NExT-QA test set. With the same Graph-SFT adapter, adding the matched graph increases accuracy from 77.86\% to 82.40\%, an improvement of 4.54 percentage points. The THG-Omni joint-stage model achieves 85.10\% without a graph and 89.54\% with the matched graph, an improvement of 4.44 percentage points. These results show that the predicted graph provides additional information for video question answering. Compared with Qwen2.5-Omni-7B, THG-Omni with the matched graph achieves an improvement of 11.56 percentage points. Fig.~\ref{fig:nextqa_case} illustrates a temporal reasoning example from the NExT-QA validation set. The question asks what the man does after the lady finishes talking. The Graph-SFT adapter predicts C without graph input and A with the matched graph, which agrees with the reference answer. The example shows how predicted graph evidence can help the model identify the relevant temporal event.

\subsection{Graph Evidence and Input Analysis}

A separate, fixed Graph-SFT adapter tests whether graph-specific training helps the model use a compact graph. With the same adapter, adding graph input raises the overall IntentBench score from 51.4805\% to 57.7330\%, a gain of 6.2525\%. Social-IQ accuracy rises from 49.6604\% to 56.6638\%, and EMER exact match from 40.6015\% to 46.6165\%. Its no-graph score is below the unadapted backbone's 55.27\%; the within-adapter benefit is therefore specific to the compact-graph condition, rather than an ablation of the final THG-Omni model. Figure~\ref{fig:social_iq_case} illustrates how one Social-IQ answer can be traced to grounded evidence. The earlier baseline also chose the reference answer in this case, so the example does not represent a corrected baseline error.

The THG-Omni rows in Table~\ref{tab:thg_main} further examine the role of graph input. The matched-graph run scores 71.72\% overall, compared with 67.77\% without graph input and 67.38\% with a shuffled graph. On Social-IQ, the corresponding accuracies are 71.18\%, 67.28\%, and 67.11\%; on MDPE, they are 68.34\%, 62.31\%, and 61.31\%. EMER sample F1 is 86.26\% with the matched graph, versus 84.63\% without a graph and 81.29\% with a shuffled graph. These observations are consistent with the model benefiting from video-matched structured evidence, particularly for social reasoning and emotion understanding. The no-graph and shuffled-graph controls were evaluated in an earlier run, however, so the differences from the updated matched-graph row should be read as comparisons of reported conditions rather than paired estimates from one evaluation run.

The three input conditions also address different questions. Removing the graph tests whether the trained model can answer from the video and available audio alone. Shuffling supplies graph text whose participant and event descriptions no longer correspond to the evaluated video; it therefore probes the importance of graph--video correspondence rather than simply the presence of additional tokens. The lower EMER score with a shuffled graph is consistent with mismatched evidence being unhelpful for emotion questions, where the relevant person and interval matter. At the same time, the no-graph condition remains substantially stronger than the unadapted backbone on Social-IQ and EMER. Part of THG-Omni's performance is therefore retained without graph input, while THG-Omni with updated matched-graph run gives the highest reported overall score. The MER24 experiment tests a separate property: when source video and audio are withheld, the predicted graph raises exact match from 14.67\% to 24.00\%. It shows that the representation contains answer-relevant information, but does not by itself measure its contribution in video-conditioned inference. Together, the controlled Graph-SFT comparison, THG-Omni input conditions, and MER24 diagnostic provide distinct evidence about learning to use a graph, preserving video correspondence, and retaining information in the graph.

The NExT-QA experiments further examine graph input on a different video question-answering dataset. On the test set, Graph-SFT achieves 82.40\% with the matched graph, compared with 77.86\% without graph input. THG-Omni further improves accuracy from 85.10\% to 89.54\% with the matched graph. These results show that graph input contributes to both graph-specific training and joint-stage adaptation. Together with the IntentBench and MER24 experiments, the NExT-QA results provide additional evidence that predicted human-centric graphs can support video reasoning.

\subsection{Annotation and Claim Verification}
\label{sec:annotation_audit}

Whole-video Qwen3-VL annotation was piloted on two development videos and then inspected on ten. All ten records passed schema validation. Manual review checked participant identity, temporal extent, label validity, and support in the video. In one case, an appearance attribute persisted across a camera transition even though the person's clothing changed; the affected interval was flagged for review. A subsequent check of 66 development videos produced 66 schema-valid records.

On the development set, HumanOmniV2 assessed 632 candidate claims from 395 valid verification units. It classified 591 as supported, 26 as unsupported, and 15 as uncertain. Only supported claims that also pass schema and participant-identity checks are eligible for semantic question construction. These verdicts are model-based assessments using the source video and available audio; the supported count is not a human-verified accuracy measure. The 632 development claims are distinct from the 632 videos in the IntentBench evaluation set.

The evaluation-video annotation run covers 632 videos. After invalid outputs are retried, all 632 annotation records pass schema validation. Schema validity checks the output format, not the truth of every semantic claim. Evaluation graphs are generated automatically: they receive no manual semantic corrections, and verifier decisions are not provided to the reasoning model at inference.

\subsection{Graph Questions and Adapter Training}
\label{sec:adapter_controls}

The graph-question collection contains 7,339 training examples: 1,555 deterministic temporal or geometric questions and 5,784 questions constructed from accepted semantic evidence. Its development collection contains 1,146 examples and has no video overlap with the graph-question training collection. Deterministic answers are computed from tracked-person statistics; semantic questions cover attributes, actions, affect, interactions, and intentions only when their source claims pass verification.

The final reasoning model is trained in two stages. First, a task question-answering LoRA adapter is trained and merged into Qwen2.5-Omni-7B. One epoch of joint training then uses the resulting checkpoint at a learning rate of $10^{-5}$. The joint epoch comprises 2,887 presentations and 181 optimizer steps: 1,155 task questions with graph dropout, 866 task questions with matched graphs, 578 verified graph questions, 144 matched-graph identification examples, and 144 mismatched-graph identification examples. These are training presentations rather than 2,887 distinct videos. LoRA updates the language-model attention projections; the visual tower, audio tower, and remaining backbone parameters are frozen during optimization. Inference uses the merged task-adapted checkpoint and the joint-stage adapter.

\subsection{Evaluation Alignment and Graph-Input Audit}
\label{sec:shuffle_audit}

IntentBench evaluation uses a frozen set of 2,688 questions and aligned predicted graphs from 632 videos: 2,356 Social-IQ, 133 EMER, and 199 MDPE questions. One video from the original 2,689-question manifest was excluded by a fixed task identifier because person tracking repeatedly exhausted GPU memory; the exclusion preceded prediction and scoring. Training-example construction does not read the held-out benchmark answers. Evaluation predictions are saved before a separate scoring step accesses reference labels.

An initial shuffled-graph control changed \texttt{structured\_evidence\_text}, while the inference formatter read \texttt{scene\_graph}. Consequently, the actual graph-text input and raw output were identical for all 2,688 questions across that control and its matched run. Those predictions are excluded from the graph-correspondence analysis. The corrected formatter reads \texttt{structured\_evidence\_text}; a pre-inference audit verified different graph-text inputs for every aligned question. Both corrected runs yielded 2,688 parsed predictions. This input audit establishes that the model received different graph text, without itself establishing a performance gain.

\subsection{Illustrative Video-to-Graph Examples}
\label{sec:illustrative_case}

Fig.~\ref{fig:app_social_iq_example} illustrates how participant identity and a temporally localized reaction can be represented alongside a Social-IQ question. The video frames and the question's reference answer come from sample \texttt{social\_iq\_001983}. The P1--P3 identifiers explain the intended representation.

Fig.~\ref{fig:app_social_iq_example2} further illustrates participant grounding and temporal visual evidence. In particular, the upper timeline organizes the selected video frames chronologically, while the visual grounding panel associates the natural-language referents with the corresponding visible participants. The temporal visual evidence panel highlights the relevant moments in the interaction, and the final panel presents the Social-IQ question together with its benchmark reference answer. Together, the two illustrations show how participant identity, temporally localized evidence, and question--answer information can be organized within the proposed video-to-graph representation.

\section{Discussion}
\label{sec:discussion}

\paragraph{Graph Construction and Graph Utilization.}
The experiments examine two related but different questions. The first is whether the predicted graph retains information useful for video reasoning. The MER24 diagnostic addresses this question by comparing question answering with and without graph input when the source video and audio are excluded. The improvement with graph input indicates that the predicted graph contains information relevant to the questions. However, this diagnostic does not measure the effect of graph input when the reasoning model also receives the source video and available audio.

The second question is whether graph-specific supervision helps the reasoning model use the predicted graph. The Graph-SFT experiments compare graph-input conditions using the same adapter. The THG-Omni experiments further examine graph input after joint training. These comparisons provide evidence about graph utilization under different training conditions. The distinction is important because a predicted graph may retain useful information even when a reasoning model does not use that information effectively. THG-Omni addresses both graph construction and graph utilization, but the two properties are evaluated separately.

\paragraph{Graph--Video Correspondence.}
The predicted graph contains participant attributes, actions, affective states, interactions, and temporal intervals associated with the evaluation video. The usefulness of this information depends on whether the graph corresponds to the correct participants and events. A shuffled graph may contain valid semantic claims from another video, but these claims do not necessarily provide relevant evidence for the current question. The shuffled-graph condition therefore examines graph--video correspondence rather than only the presence of additional graph text.

The IntentBench results are consistent with matched graph input providing more useful information than shuffled graph input. However, the updated matched-graph condition and earlier control conditions were not all evaluated in one paired run. The reported differences should therefore be interpreted as comparisons between the available evaluation conditions rather than paired statistical estimates. The corrected graph-input audit establishes that the matched and shuffled conditions receive different graph text, but it does not independently establish a performance improvement.

\paragraph{Limitations of Predicted Graphs.}
The quality of the predicted graph depends on person tracking, semantic annotation, and identity binding. SAM3 provides person tracks and geometric relations, but occlusion, camera transitions, and changes in participant appearance may affect the available geometric evidence. Qwen3-VL provides semantic claims, but these claims may contain incorrect participant associations or temporal intervals. Although human quality control and HumanOmniV2 verification are used during training-data construction, these procedures cannot guarantee the correctness of every retained claim.

Evaluation graphs introduce an additional limitation. They are generated automatically from evaluation videos without manual semantic correction or claim verification. Consequently, a predicted graph may omit relevant information or include semantic claims that are not fully supported by the source video. The reasoning model must use the predicted graph together with the source video and available audio under these conditions. The compact-graph validation results also indicate that including more semantic claims does not necessarily lead to higher accuracy. The effect of graph content may depend on the question and the information retained in the graph.

\paragraph{Future Work.}
Future work will focus on improving participant identity binding and the reliability of semantic claims in predicted graphs. More reliable person tracking across camera transitions may reduce incorrect associations between semantic claims and person tracks. Further work may also examine how to select graph fields according to the question, so that the reasoning model receives relevant participant and temporal evidence without unnecessary semantic claims. In addition, evaluating THG-Omni on longer videos and more complex interactions would help examine the effect of graph construction and graph-specific supervision under different video conditions.

\section{Conclusion}
In this work, we present THG-Omni, a framework for human-centric video reasoning with traceable graph supervision. It links SAM~3 participant tracks with temporally localized semantic annotations, verifies candidate claims against the source evidence, and uses the resulting graph to train an omni-modal model to interpret people, actions, affect, and interactions. Experiments on IntentBench show consistent improvements over the unadapted backbone across social reasoning, emotion understanding, and deception perception. Experiments on NExT-QA further examine the use of predicted graph evidence for causal, temporal, and descriptive video question answering. Additional graph-input experiments show that the structured representation retains answer-relevant evidence and can improve reasoning when the model is trained to use it. THG-Omni thus offers a way to improve video question answering while making the supporting people and events easier to identify.

\thispagestyle{empty}

\section*{Author Biography}
\authorbibliography[scale=0.09,overhang=0pt,wraplines=9,imagewidth=3cm,imagepos=l]{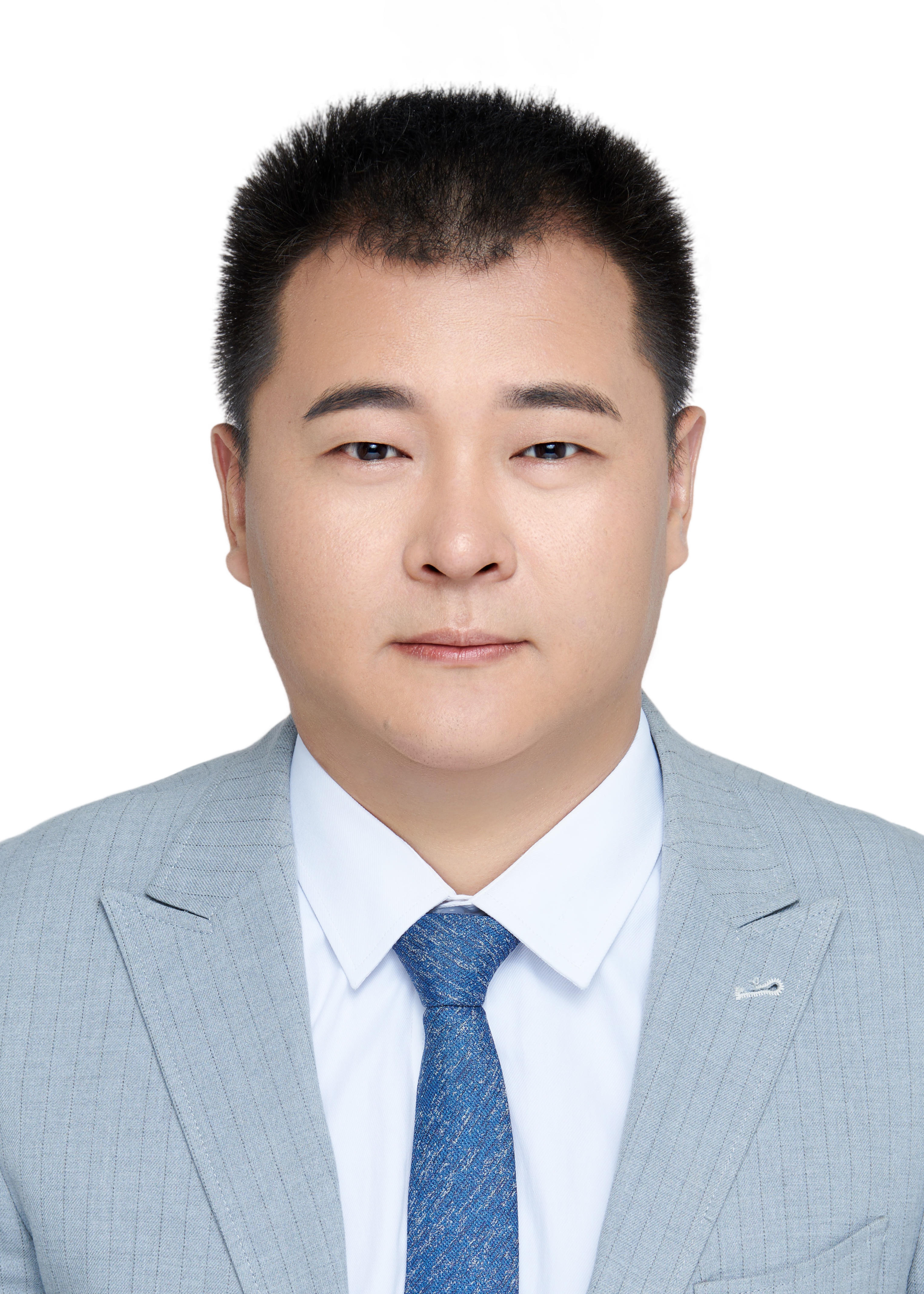}{Jisheng Dang \rm{
received the Ph.D. degree from Sun Yat-sen University, China, advised by Prof. Jianhuang Lai and Prof. Huicheng Zheng.
He worked as a research fellow at the NExT++ laboratory of the National University of Singapore, advised by Prof. Tat-Seng Chua. 
He is now a tenured associate professor at the School of Information Science and Engineering, Lanzhou University. 
His research interests include multimodal learning, video understanding, and embodied intelligence. 
He has published several papers as the first author or corresponding author in major journals and conferences including IEEE TPAMI/TIP/TAC/TNNLS/ECCV/TITS/AAAI/ICLR/NeurIPS.
}}

\authorbibliography[scale=0.8,overhang=0pt,wraplines=11,imagewidth=3cm,imagepos=l]{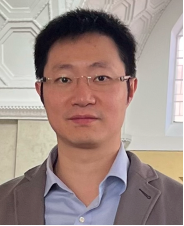}{Wei-Shi Zheng \rm{
is a full Professor with Sun Yat-sen University. 
His research interests include person association, activity understanding, AI robotics learning, and related weakly supervised, unsupervised, and continuous learning algorithms. 
He has published more than 200 papers, including over 150 in leading journals (TPAMI, IJCV, SIGGRAPH, TIP) and top conferences (ICCV, CVPR, ECCV, NeurIPS). 
He has served as Area Chair for ICCV, CVPR, ECCV, BMVC, and NeurIPS.  
He is a Cheung Kong Scholar Distinguished Professor, recipient of the NSFC Excellent Young Scientists Fund, a Fellow of IAPR, and recipient of the Royal Society ewton Advanced Fellowship (UK). He is an Associate Editor and editorial board member of IEEE TPAMI, Artificial Intelligence Journal, and Pattern Recognition.
}}

\authorbibliography[scale=0.115,overhang=0pt,wraplines=11,imagewidth=3cm,imagepos=l]{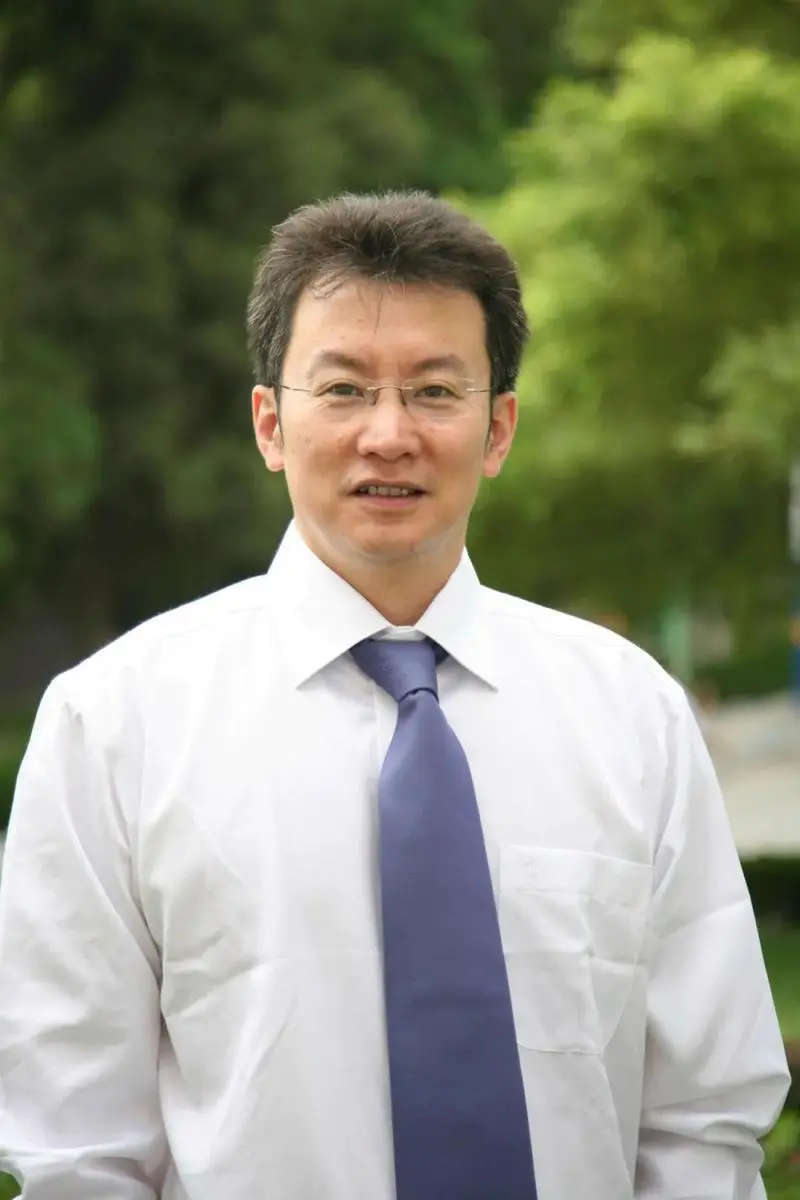}{Bin Hu (Fellow, IEEE) \rm{
received the Ph.D. degree in computer science from the Institute of Computing Technology, Chinese Academy of Sciences, China, in 1998. 
Since 2008, he has been a Professor and Dean of the School of Information Science and Engineering, Lanzhou University. 
He has also held a guest professorship at ETH Zurich. 
He serves as Editor-in-Chief of IEEE Transactions on Computational Social Systems and is a Fellow of IET and AAIA. 
His research interests include pervasive computing, computational psychophysiology, data modeling, and artificial intelligence.
}}

\authorbibliography[scale=0.69,overhang=0pt,wraplines=11,imagewidth=3cm,imagepos=l]{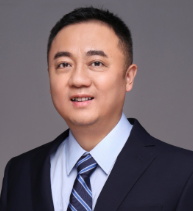}{Qi Tian (Fellow, IEEE) \rm{ received the Ph.D. degree in ECE from the University of Illinois at Urbana-Champaign, Champaign, IL, USA, in 2002. He is currently the Chief Scientist of Huawei Terminal BG. He was the Chief Scientist in computer vision with Huawei Noah's Ark Laboratory from 2018 to 2020. Before he joined Huawei, he was a Full Professor with the Department of Computer Science, The University of Texas at San Antonio, San Antonio, TX, USA, from 2002 to 2019. He was listed in the top ten of the 2016 Most Influential Scholars in Multimedia by Aminer.org. He is an IEEE Fellow (Class 2016) and an Academician of the International Eurasian Academy of Sciences (elected in 2021), with  113,457   citations on Google Scholar.
}}
 
\authorbibliography[scale=0.12,overhang=0pt,wraplines=11,imagewidth=3cm,imagepos=l]{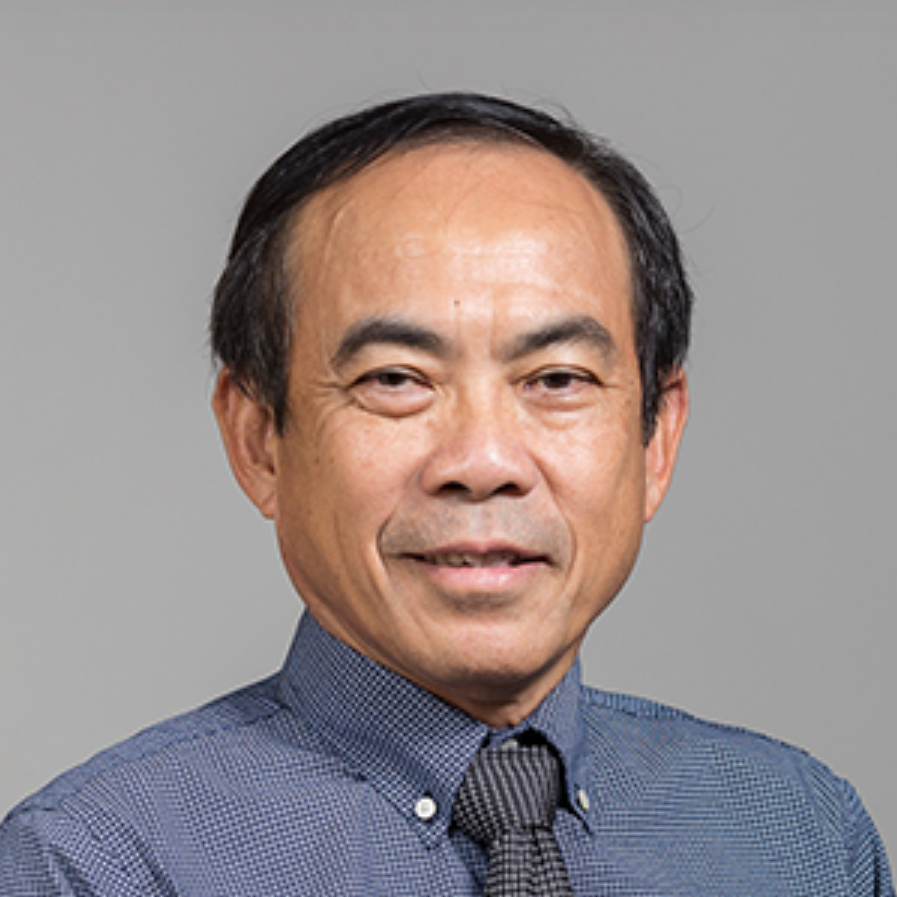}{Tat-Seng Chua \rm{received the Ph.D. degree from the University of Leeds, U.K. He is the KITHCT chair professor with the School of Computing, National University of Singapore, where he was the acting and founding dean of the School from 1998 to 2000. He is the co-director of NExT, a joint center between NUS and Tsinghua University, to develop technologies for live social media search. He is the 2015 winner of the prestigious ACM SIGMM Award. He is the chair of the Steering Committee of the ACM International Conference on Multimedia Retrieval (ICMR) and the Multimedia Modeling (MMM) conference series.   He is also the general co-chair of ACM Multimedia 2005, ACM CIVR (now ACM ICMR) 2005, ACM SIGIR 2008, and ACM Web Science 2015.   He serves on the editorial boards of four international journals. He is the co-founder of two technology startups in Singapore and a Fellow of the Singapore Academy of Sciences, with  114,163  citations on Google Scholar.}}

\vfill

\end{document}